\documentclass{article}

\usepackage{arxiv}

\RequirePackage[table]{xcolor}
\usepackage{array,hhline}
\usepackage{pgf}
\newcommand*{\gray}{gray}
\newcommand*{\woB}[1]{\multicolumn{1}{c}{#1}}

\newcommand*{\colorme}[1]{%
    \pgfmathparse{#1<.5?1:0}%
    \ifnum\pgfmathresult=0\relax\color{white}\fi
    \pgfmathparse{1-#1}
    \expandafter\cellcolor\expandafter[\expandafter\gray\expandafter]\expandafter{\pgfmathresult}%
    #1%
}
\newlength{\tabwidth}
\usepackage[utf8]{inputenc} 
\usepackage[T1]{fontenc}    
\usepackage{hyperref}       
\usepackage{url}            
\usepackage{booktabs}       
\usepackage{amsfonts}       
\usepackage{nicefrac}       
\usepackage{microtype}      
\usepackage{lipsum}
\graphicspath{ {./images/} }

\title{Control of computer pointer using hand gesture recognition in motion pictures}

\author{
 Yalda Foroutan \\
  School of Electrical and Computer Engineering\\
  University of Tehran\\
  \texttt{yalda.foroutan@gmail.com} \\
   \And
 Ahmad Kalhor \\
  School of Electrical and Computer Engineering\\
  University of Tehran\\
  \texttt{akalhor@ut.ac.ir} \\
  \And
 Saeid Mohammadi nejati \\
  School of Electrical and Computer Engineering\\
  University of Tehran\\
  \texttt{Saeid.nejati.moh@ut.ac.ir} \\
  \And
 Samad Sheikhaei \\
  School of Electrical and Computer Engineering\\
  University of Tehran\\
  \texttt{sheikhaei@ut.ac.ir} \\
}

\begin{document}
\maketitle

\begin{abstract}
This paper presents a user interface designed to enable computer cursor control through hand detection and gesture classification. A comprehensive hand dataset comprising 6720 image samples was collected, encompassing four distinct classes: fist, palm, pointing to the left, and pointing to the right. The images were captured from 15 individuals in various settings, including simple backgrounds with different perspectives and lighting conditions. A convolutional neural network (CNN) was trained on this dataset to accurately predict labels for each captured image and measure their similarity. The system incorporates defined commands for cursor movement, left-click, and right-click actions. Experimental results indicate that the proposed algorithm achieves a remarkable accuracy of 91.88\% and demonstrates its potential applicability across diverse backgrounds.
\end{abstract}


\section{Introduction}
In today's world, computers play an indispensable role in various aspects of human life, permeating both personal and social spheres. The widespread availability and mass production of personal computers have had a profound impact on our daily lives. Human-Computer Interaction (HCI) aims to enhance and facilitate communication between humans and computers by creating interfaces that can interpret hand gestures as meaningful commands. Despite the advancements in HCI, traditional input devices such as computer mice are still widely used. However, the mouse, which was introduced over 50 years ago, has undergone significant improvements over time. Despite these improvements, the mouse relies on direct physical contact and restricts the user to controlling the computer from a close proximity.

Moreover, with the advent of the COVID-19 pandemic, there has been a growing need for alternative input methods that do not rely on physical contact with devices. This situation has accelerated the exploration and development of contactless interfaces for controlling machines and computers. By leveraging hand gestures, a user can interact with a computer from a distance, reducing the risk of transmission and ensuring a safer computing experience.

This paper proposes a novel approach to human-computer interaction through hand gesture recognition. The objective is to design a user interface that enables users to control the computer cursor using hand gestures, eliminating the need for physical contact or reliance on traditional input devices. The system employs a dataset of hand images captured from a diverse group of individuals, encompassing different hand gestures such as fist, palm, pointing to the left, and pointing to the right.

The remainder of this paper is organized as follows: Section 2 provides an overview of related work in hand gesture recognition. Section 3 describes the dataset collection process and its composition. Section 4 presents the methodology employed, including the architecture and training of the CNN model. Section 5 discusses the experimental results and evaluates the performance of the proposed system. Finally, Section 6 concludes the paper, highlighting its contributions and discussing potential avenues for future research and development.

\section{Literature review}
In recent years, there has been growing interest among machine learning enthusiasts in studying human activities as a means to explore alternative methods for computer control~\cite{gil2020human}. Several studies have focused on using different approaches to control computers, such as utilizing Kinect sensors~\cite{ren2013robust}, EEG mice~\cite{alomari2014eeg}, or EMG signals~\cite{benalcazar2017real} to classify human actions and assign corresponding mouse commands, thereby controlling the computer pointer. However, these methods often rely on additional hardware that is more expensive and bulkier than traditional computer mice. To overcome these limitations, software-based solutions for controlling the computer pointer offer a more promising approach.

Given that hands are the most commonly used body part for manipulating objects, leveraging hand gestures to control computer systems presents an effective way to replace traditional mouse hardware. Previous works on hand gesture recognition have employed both contact-based and contactless approaches. Older contact-based methods involved using data gloves~\cite{kim20093} to detect hand gestures and track their movements. However, these gloves were often cumbersome, hindering natural hand movements due to their weight, wiring, and sensor placements. With the advancements in computer vision, newer iterations of data gloves have become simpler, eliminating wiring and relying on camera-based hand tracking techniques~\cite{wang2009real}. In some cases, data gloves have been replaced by colored fingertips~\cite{mistry2009sixthsense}. Eventually, hand gesture applications have evolved into touchless systems, utilizing machine learning techniques and captured frames from cameras.

Machine learning techniques that utilize image processing systems, such as cameras, offer an alternative to contact-based approaches. These techniques provide greater freedom of movement for hands, a crucial aspect for effective Human-Computer Interaction (HCI) systems. Image processing methods are used to convert camera-captured images into digital forms, enabling scaling, filtering, and noise removal. Computer vision algorithms then enable computers to discern between different gestures, mimicking human visual perception. For example, in~\cite{veluchamy2015vision}, skin color detection is employed by setting a color threshold to detect human skin and remove the background from images. Another approach in~\cite{dhule2014computer} involves analyzing video sequences instead of individual frames, utilizing methods such as skin detection and approximate median models. ~\cite{huang2019hand} detects both hands and heads based on skin color, creating binary masks for each and training a VGGNet model to distinguish between them. In~\cite{park2008method}, hand gestures are discriminated based on the angle between the thumb and index finger. Other papers, such as~\cite{noreen2015hand}, convert images to the HSV color space to access color information efficiently. In~\cite{grif2018human}, hands are captured against a blue background to take advantage of the distinct variations between human skin and the background in the HSV space.

Hand gesture recognition typically involves two stages: hand localization and gesture classification. In the past, before the emergence of neural networks, SIFT algorithm was widely used for feature extraction due to its computational efficiency. For instance, Golash and Kulkarni~\cite{golash2020economical} utilized SIFT to distinguish between eight gestures for controlling machines like fans and washing machines. However, color-based methods, although simple and easy to implement, often suffer from limited generalization across different individuals or challenging conditions. These methods often exhibit dependencies on skin color, as users are required to manually input their specific skin color, as demonstrated in~\cite{globefire}. Furthermore, pixel-wise differentiation between human skin and backgrounds is more complex and sensitive than it may initially appear.

The availability of large datasets has paved the way for the use of neural networks as a substitute for traditional machine learning approaches, offering greater robustness to light conditions, perspective variations, and diverse backgrounds. Neural network-based object detection algorithms can be employed for hand localization. Algorithms such as You Only Look Once (YOLO) and Single Shot Multi-Box Detector (SSD) are particularly suitable for real-time tasks and provide higher accuracy compared to R-CNN or Faster R-CNN. ~\cite{yi2018long} utilizes two SSD detectors: the first detects the head and shoulder area, while the second recognizes hand gestures within the detected region. The SSD algorithm can be fine-tuned with new images\cite{liu2019hand}, and selective dropout techniques can reduce computational load~\cite{ni2018light} to minimize performance lag. By merging hand detection and gesture recognition during the detection phase, a valid predicted label can be obtained~\cite{tanmaiehand}. Deep learning-based hand gesture recognition algorithms have demonstrated superior accuracy compared to traditional methods. While deep learning approaches eliminate the need for manual feature extraction and reduce reliance on handcrafted designs, they can be computationally intensive. For instance, ~\cite{haratiannejadi2019smart} combines a data glove with hand gesture recognition using an SSD-based hand detector and an SVM for gesture classification, enabling the recognition of gestures from both hands.

\section{Dataset}
\begin{figure}[ht]
	\centerline{\includegraphics[width=0.4\textwidth]{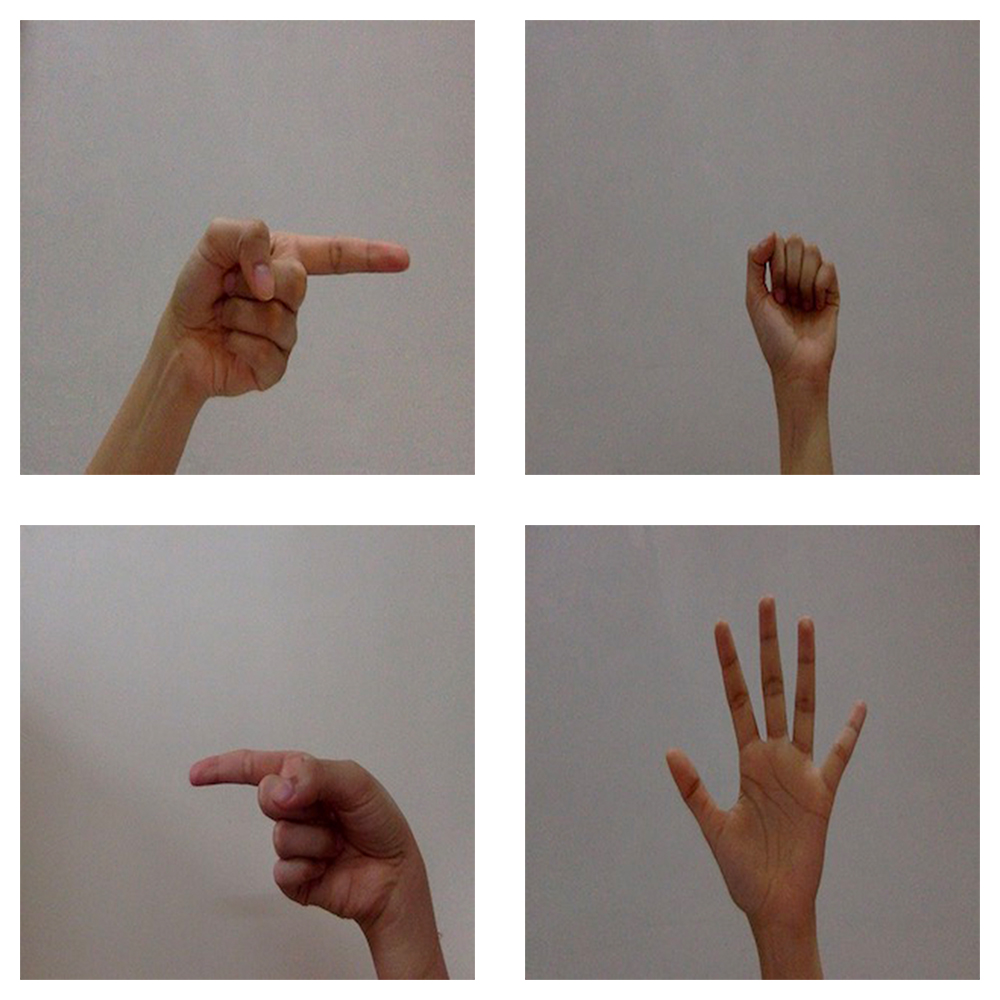}}
    \caption{Examples from the collected hand dataset. The samples were captured under various conditions, including different backgrounds, lighting conditions, and distances from webcams.}
    \label{fig:dataset}
\end{figure}

A hand dataset consisting of 6720 image samples (300 x 300 pixels) was collected for this study. The dataset includes images of 15 subjects against 18 different simple backgrounds. The dataset encompasses four distinct gesture classes: fist, palm, pointing to the right, and pointing to the left. Participants were instructed to capture images of both hands, showcasing both the palmar and dorsal sides. Four samples from the dataset, representing each gesture class, are illustrated in Figure \ref{fig:dataset}.

For the dataset split, 5120 samples were allocated for the training set, while 1600 samples were evenly divided between the validation and test sets. To ensure diverse distributions and mitigate overfitting, the training and validation sets were intentionally constructed with different distributions. The training samples were captured directly from webcams without any intermediary software, while the validation samples were selected based on the acceptable images obtained from the SSD algorithm used for hand detection. Consequently, the CNN classifier was trained on one distribution and evaluated on another distribution for the validation data and real-time scenarios.

\section{Proposed Algorithm}
\begin{figure}[h!]
	\centerline{\includegraphics[width=1\textwidth]{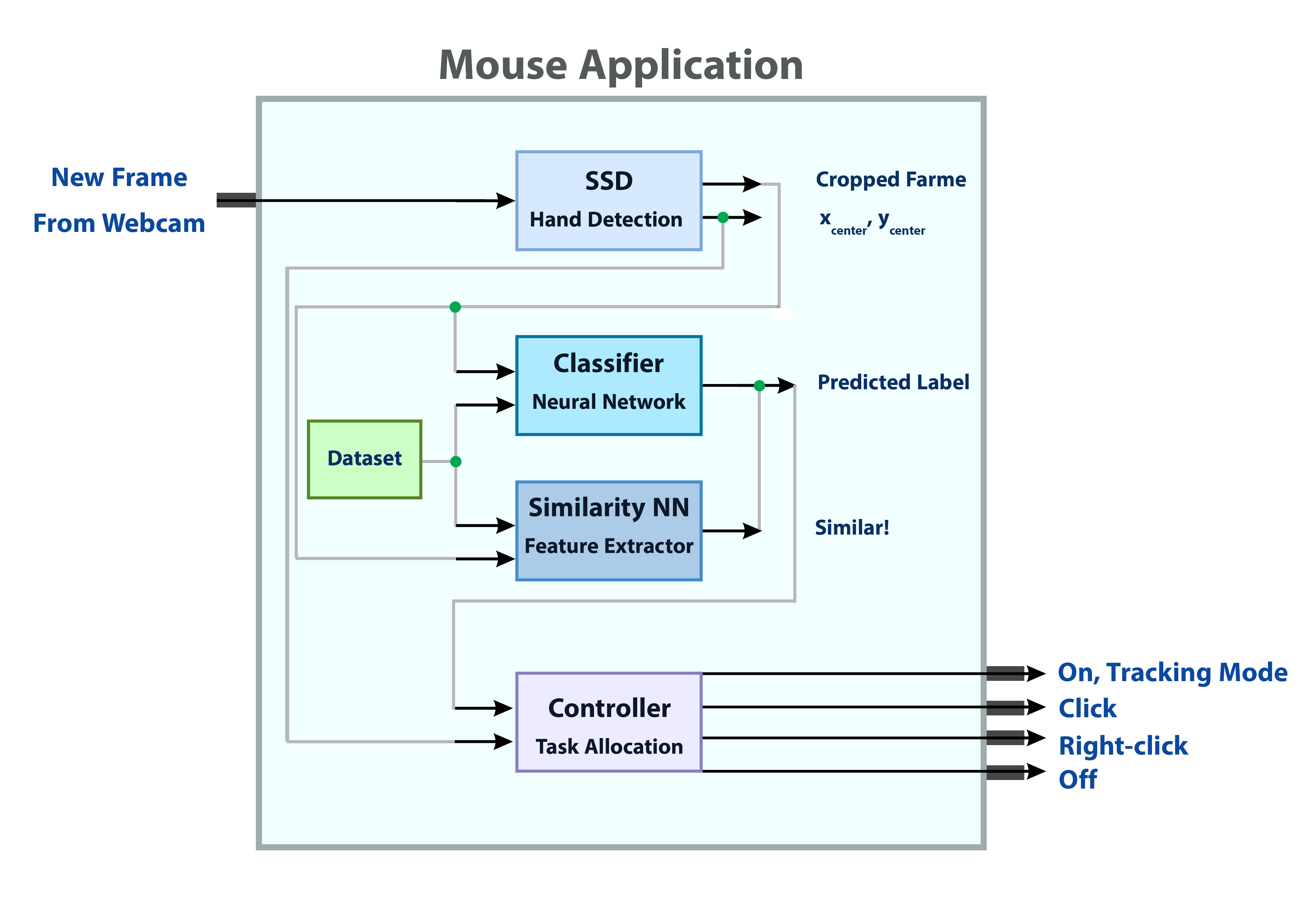}}
    \caption{The proposed algorithm for controlling the computer cursor.}
    \label{fig:algorithm}
\end{figure}

In this section, a human-computer interface (HCI) based on hand gestures for controlling the computer's pointer is designed. The algorithm consists of several components, including hand detection, classification, and mouse commands. The overall flow of the algorithm is summarized in Figure \ref{fig:algorithm}.

\subsection{Hand Detection}
\begin{figure}[ht]
	\centerline{\includegraphics[width=0.9\textwidth]{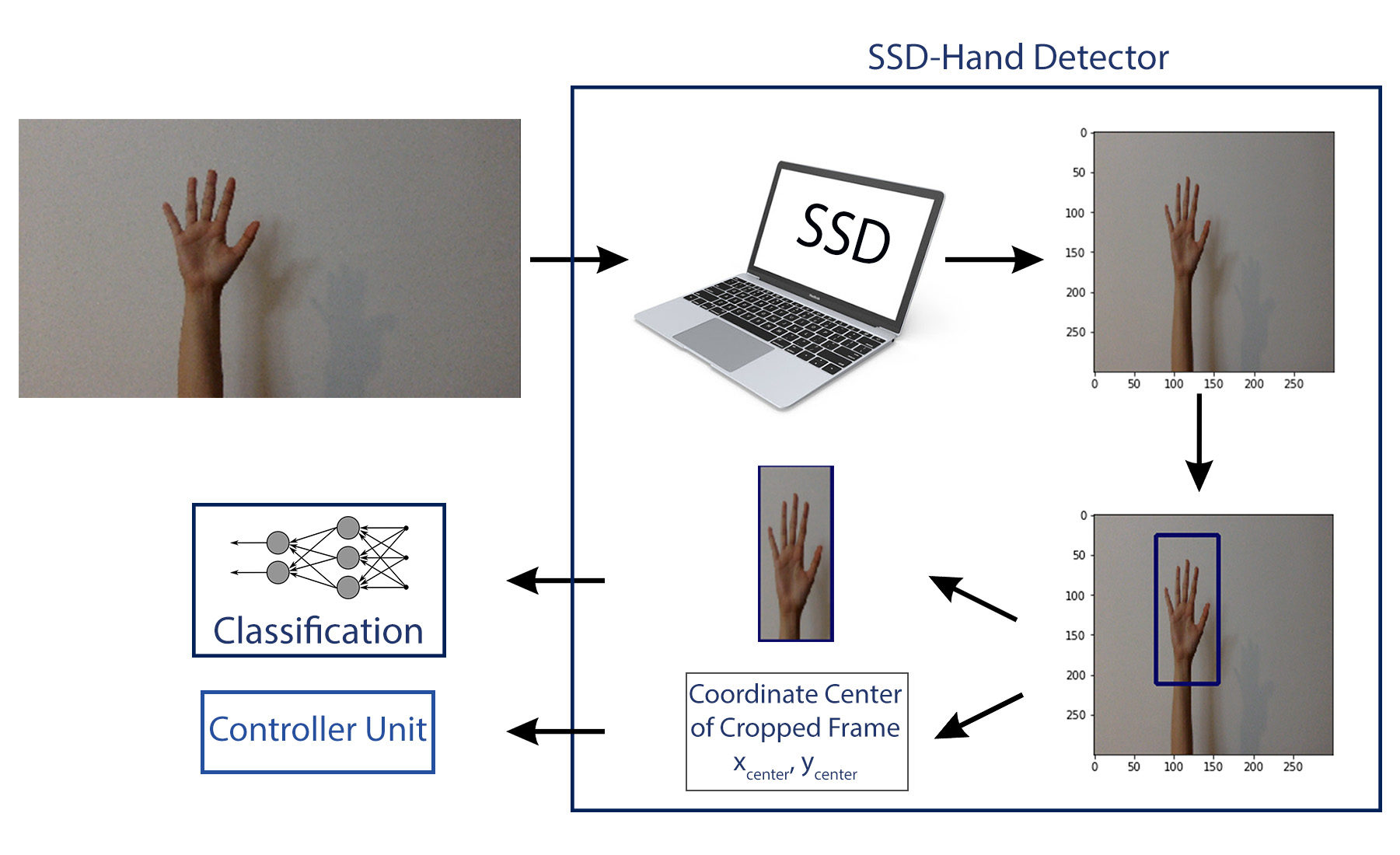}}
    \caption{The operation of the SSD hand detector. A captured frame is passed through the SSD hand detector, resulting in a cropped frame containing the hand region and the center coordinate of the cropped frame.}
    \label{fig:ssd}
\end{figure}

The algorithm starts by capturing frames from the computer's webcam. These frames are then pre-processed and passed through a hand detection algorithm based on Single Shot Multi-Box Detector (SSD). The SSD algorithm is capable of detecting hands in the frames and provides two outputs: a cropped frame containing the hand region and the center coordinate of the cropped frame. If a hand is detected in the frame, the algorithm proceeds to the next step. Otherwise, it continues to the next frame for hand detection. The process of hand detection is illustrated in Figure \ref{fig:ssd}.

\subsection{Classification}
The cropped frames containing the hand regions are then fed into the classification part of the algorithm. Since the SSD detector can detect various hand gestures, the cropped frames can belong to one of the four defined classes in the dataset or represent undefined gestures. The goal of the classification part is to predict a valid label for frames with defined classes and ignore frames with undefined gestures.

To achieve this, a Convolutional Neural Network (CNN) is trained using the EfficientNet-B0 architecture followed by eight fully connected layers for classification. The last layer of the EfficientNet-B0 network, with 1280 neurons, is connected to the fully connected layers, which ultimately reduces the output to four neurons corresponding to the four defined gesture classes. The CNN is trained using the collected hand samples from the dataset, which have dimensions of 70 x 70 x 3, for 20 epochs. The trained CNN achieves 99 percent accuracy on the test set.

To remove unwanted classes and ensure valid predictions, a Radial Basis Function (RBF) network is designed. The last eight dense layers of the trained CNN are removed to create a similarity network. This similarity network acts as an encoder or feature extractor, reducing the dimensionality of the samples from 70 x 70 x 3 to 1280. Mean vectors are calculated for each class by feeding the encoded samples of each class through the network. These mean vectors serve as reference vectors.

During the evaluation phase, the cropped frames from the validation and test sets are preprocessed and fed into the frozen similarity network. The encoded samples are then compared with the reference vectors using Euclidean distance. A threshold is defined for each class based on the maximum distance within that class. When a cropped frame is input to the classification part, its output is compared with the reference vectors, and the smallest distance is chosen. If the chosen distance is below its corresponding threshold, the cropped frame is considered to belong to one of the four defined classes. Otherwise, it is classified as an unwanted class and should be ignored, prompting the algorithm to process the next frame in the hand detection stage.

Therefore, the classification part performs two tasks: the classifier predicts a label for the cropped frame, and the similarity network compares the cropped frame with the reference vectors and thresholds to determine whether it represents one of the four dataset classes or not. The classifier and similarity network act independently, and the result from the similarity network validates the predicted label. If the classification part predicts a valid label, the computer cursor will respond accordingly.

\begin{figure}[h!]
	\centerline{\includegraphics[width=0.7\textwidth]{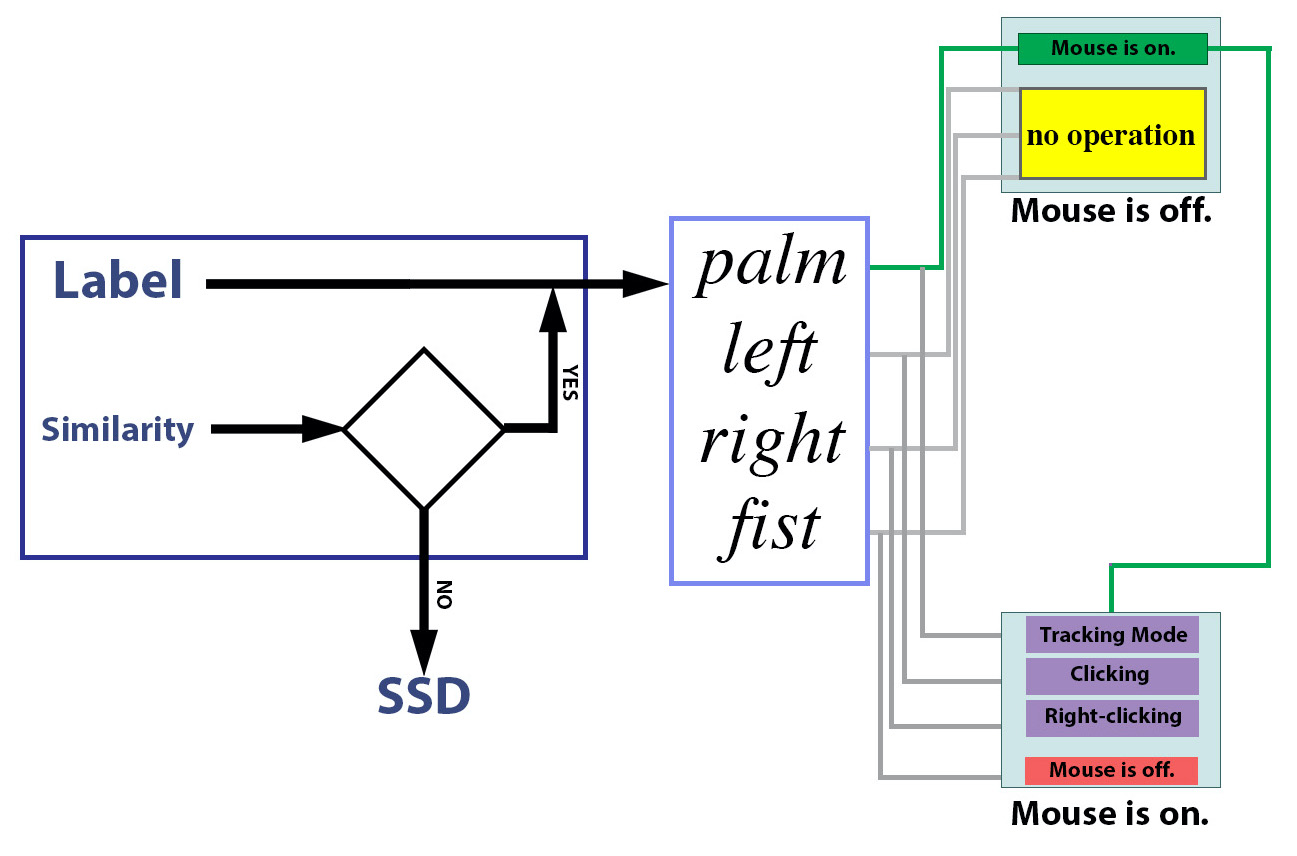}}
    \caption{The designed control unit for the HCI. The cursor can be moved, clicked, or right-clicked based on the valid label received from the classification part.}
    \label{fig:controller}
\end{figure}

\subsection{Mouse commands}
To control the computer's pointer based on the predicted labels, a controller unit is designed. Initially, the proposed algorithm is turned off, and it is activated by detecting the user's palm. Once the palm is detected, the algorithm enables the recognition of hand gestures and allows the user to control the cursor.

The recognized palm movements are used to move the cursor based on the center coordinate of the cropped frame. Since the input frames for the SSD algorithm have a resolution of 300 x 300, the coordinates need to be converted to meaningful screen coordinates. When the algorithm is activated, the user can perform left-pointing and right-pointing gestures to initiate left-click and right-click actions, respectively. To deactivate the algorithm, the user can make a fist gesture, and no action will occur until the palm is detected again to activate the algorithm. Figure \ref{fig:controller} illustrates the control unit design.

\begin{figure}[h!]
	\centerline{\includegraphics[width=1\textwidth]{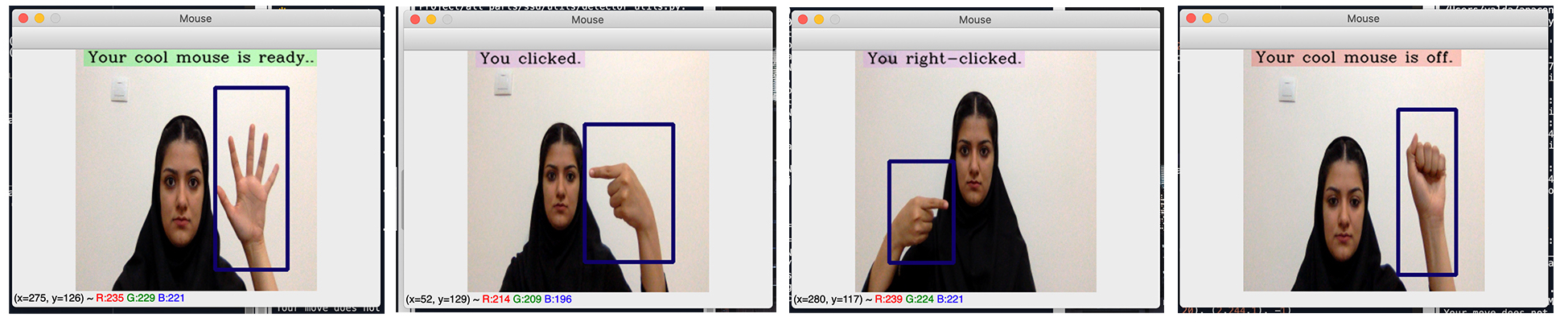}}
    \caption{Controlling the cursor of the computer through hand gesture recognition. The algorithm is activated by recognizing the user's palm, and the cursor can be moved by the center coordinate of the cropped frame. Left-pointing and right-pointing gestures enable left-click and right-click actions, respectively. The algorithm is deactivated by recognizing the user's fist.}
    \label{fig:mouse}
\end{figure}

The overall process of controlling the computer cursor through hand gesture recognition is depicted in Figure \ref{fig:mouse}.

The proposed algorithm provides a robust and intuitive method for human-computer interaction using hand gestures. By leveraging hand detection, gesture classification, and mouse commands, users can control the computer's pointer in a natural and efficient manner.

\section{Experimental Results}
The proposed algorithm achieved a high classification accuracy of 99\% using the EfficientNet-B0 architecture. Compared to VGG16, EfficientNet-B0 had significantly fewer parameters while maintaining a comparable accuracy. This indicates that EfficientNet-B0 is a more efficient and lightweight architecture for the hand gesture classification task.

The algorithm's performance was evaluated using different backgrounds, distances from the webcam, and light conditions. Overall, the algorithm demonstrated good accuracy across various scenarios. However, the clicking mode showed slightly lower accuracy, especially in the presence of a complex background. The confusion matrix revealed some mis-classifications between the clicking and turn-off modes in such cases. This is likely due to the SSD hand detector misidentifying the hand gesture in the complex background.

The algorithm's speed of 15 frames per second allows for real-time control of the computer pointer, enabling smooth and responsive interaction. This is an important aspect for user experience, as a slow response can hinder the usability of the system.

Overall, the proposed algorithm demonstrates the feasibility and effectiveness of using hand gestures for controlling the computer's pointer. With further improvements in the hand detection stage, such as using advanced algorithms or additional preprocessing techniques, the accuracy of the clicking mode can be enhanced, resulting in a more robust and reliable system.

\subsection{Limitations and Future Work}

While the proposed algorithm shows promising results, there are some limitations and areas for future improvement:

\begin{itemize}
\item \textbf{Complex Backgrounds}: The algorithm's performance was slightly affected by complex backgrounds, leading to mis-classifications, particularly in the clicking mode. Improving the hand detection algorithm or incorporating background subtraction techniques can help mitigate this issue.
\item \textbf{Hand Occlusion}: The algorithm assumes that both hands are visible and does not handle occlusion or partial visibility of the hands. Developing techniques to handle occlusion can enhance the algorithm's robustness in real-world scenarios.
\item \textbf{Gesture Expansion}: The algorithm currently supports four defined hand gestures. Expanding the dataset and training the model on additional hand gestures can enable a wider range of functionalities and control options.
\item \textbf{Dynamic Hand Gestures}: The algorithm focuses on static hand gestures. Incorporating dynamic hand gestures and tracking hand movements can provide more nuanced control and enable additional interactions.
\item \textbf{User Adaptation}: The algorithm does not adapt to individual users. Personalizing the algorithm based on user-specific hand gestures and preferences can improve accuracy and user satisfaction.
\end{itemize}

Addressing these limitations and conducting further user studies can lead to a more refined and user-friendly hand gesture-based control system for human-computer interaction.

\begin{figure}[ht]
	\centerline{\includegraphics[width=1\textwidth]{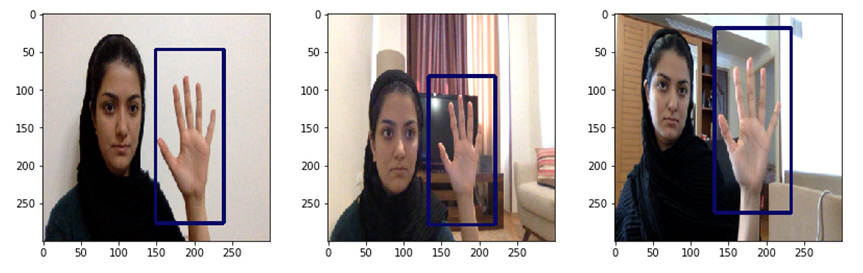}}
    \caption{Selected backgrounds for algorithm evaluation.}
    \label{fig:backgrounds}
\end{figure}

As mentioned earlier, the training set has a distinct distribution compared to the validation and test sets. During the learning process, approximately 76\% of the dataset is used for training, while the remaining samples are allocated for validation and testing. To evaluate the proposed algorithm, various scenarios were considered, including three new backgrounds (white, simple, and complex), two distances from the webcam, and two lighting conditions. In each scenario, ten frames containing both hands were captured. The three selected backgrounds can be seen in Figure \ref{fig:backgrounds}. Thus, a total of 80 frames were examined for each hand position, and the results are presented in Table \ref{table:acc}.

The confusion matrix of the proposed algorithm is provided in Table \ref{table:Confusion_matrix}. The clicking mode exhibits the lowest accuracy due to the presence of a brown closet in the complex background, which caused confusion for the SSD algorithm in accurately detecting hand gestures.

\begin{table}[ht]
\caption{Performance of the classification part with different architectures}
\centering
\begin{tabular}{c c c c}
\hline\hline
Network & Accuracy & Parameters & Run-time \\ [0.5ex] 
\hline
EfficientNet-B0 &99\%   &4.98M  &671us/step \\
VGG16           &100\%  &15.95M &776us/step \\[1ex]
\hline
\end{tabular}
\label{table:acc}
\end{table}

\begin{table}
\centering
\caption{Confusion matrix for each mode of controlling the cursor}
\begin{tabular}{c c c c c}
    \woB{} & \woB{Turn Off}         & \woB{Turn On}      & \woB{Click}        & \woB{R-click}         \\ \hhline{~*4{|-}|}
     Turn Off & \colorme{0.9125}    & 0                  & \colorme{0.1375}   & \colorme{0.0375}      \\ \hhline{~*4{|-}|}
     Turn On  & \colorme{0.0750}    & \colorme{0.9708}   & 0                  & 0                     \\ \hhline{~*4{|-}|}
     Click & \colorme{0.0083}       & \colorme{0.0250}   & \colorme{0.8625}   & \colorme{0.0333}      \\ \hhline{~*4{|-}|}
     R-click & \colorme{0.0041}     & \colorme{0.0041}   & 0                  & \colorme{0.9292}      \\ \hhline{~*4{|-}|}
\end{tabular}
\label{table:Confusion_matrix}
\end{table}

\section{Conclusion}
In conclusion, this paper presented a touch-less human-computer interaction algorithm based on hand gesture recognition for controlling the computer pointer. The proposed algorithm utilized deep learning methods, particularly convolutional neural networks (CNNs), to achieve accurate gesture recognition.

The algorithm demonstrated its effectiveness in complex backgrounds, different lighting conditions, and varying camera perspectives. It offers a valuable solution for intelligent systems and can be particularly useful in public places due to its touch-less nature, which is especially relevant in the context of the COVID-19 virus.

The designed computer cursor controller exhibited promising results, achieving an accuracy of 91.88\% for various backgrounds and 97\% for simpler environments. Additionally, a hand gesture dataset comprising 6720 colored images with four classes was introduced, which can be utilized for various other purposes beyond this particular algorithm.

Overall, this touch-less human-computer interaction algorithm opens up possibilities for enhanced interaction with computer systems and has potential applications in a wide range of domains.

\bibliographystyle{unsrt}  

\begin{thebibliography}{10}

\bibitem{gil2020human}
Manuel Gil-Mart{\'\i}n, Rub{\'e}n San-Segundo, Fernando
  Fern{\'a}ndez-Mart{\'\i}nez, and Ricardo de~C{\'o}rdoba.
\newblock Human activity recognition adapted to the type of movement.
\newblock {\em Computers \& Electrical Engineering}, 88:106822, 2020.

\bibitem{ren2013robust}
Zhou Ren, Junsong Yuan, Jingjing Meng, and Zhengyou Zhang.
\newblock Robust part-based hand gesture recognition using kinect sensor.
\newblock {\em IEEE transactions on multimedia}, 15(5):1110--1120, 2013.

\bibitem{alomari2014eeg}
Mohammad~H Alomari, Ayman AbuBaker, Aiman Turani, Ali~M Baniyounes, and Adnan
  Manasreh.
\newblock Eeg mouse: A machine learning-based brain computer interface.
\newblock {\em Int. J. Adv. Comput. Sci. Appl}, 5(4):193--198, 2014.

\bibitem{benalcazar2017real}
Marco~E Benalc{\'a}zar, Cristhian Motoche, Jonathan~A Zea, Andr{\'e}s~G
  Jaramillo, Carlos~E Anchundia, Patricio Zambrano, Marco Segura,
  Freddy~Benalc{\'a}zar Palacios, and Mar{\'\i}a P{\'e}rez.
\newblock Real-time hand gesture recognition using the myo armband and muscle
  activity detection.
\newblock In {\em 2017 IEEE Second Ecuador Technical Chapters Meeting (ETCM)},
  pages 1--6. IEEE, 2017.

\bibitem{kim20093}
Ji-Hwan Kim, Nguyen~Duc Thang, and Tae-Seong Kim.
\newblock 3-d hand motion tracking and gesture recognition using a data glove.
\newblock In {\em 2009 IEEE International Symposium on Industrial Electronics},
  pages 1013--1018. IEEE, 2009.

\bibitem{wang2009real}
Robert~Y Wang and Jovan Popovi{\'c}.
\newblock Real-time hand-tracking with a color glove.
\newblock {\em ACM transactions on graphics (TOG)}, 28(3):1--8, 2009.

\bibitem{mistry2009sixthsense}
Pranav Mistry and Pattie Maes.
\newblock Sixthsense: a wearable gestural interface.
\newblock In {\em ACM SIGGRAPH ASIA 2009 Art Gallery \& Emerging Technologies:
  Adaptation}, pages 85--85. 2009.

\bibitem{veluchamy2015vision}
S~Veluchamy, LR~Karlmarx, and J~Jeya Sudha.
\newblock Vision based gesturally controllable human computer interaction
  system.
\newblock In {\em 2015 International Conference on Smart Technologies and
  Management for Computing, Communication, Controls, Energy and Materials
  (ICSTM)}, pages 8--15. IEEE, 2015.

\bibitem{dhule2014computer}
Chetan Dhule and Trupti Nagrare.
\newblock Computer vision based human-computer interaction using color
  detection techniques.
\newblock In {\em 2014 Fourth International Conference on Communication Systems
  and Network Technologies}, pages 934--938. IEEE, 2014.

\bibitem{huang2019hand}
Hanwen Huang, Yanwen Chong, Congchong Nie, and Shaoming Pan.
\newblock Hand gesture recognition with skin detection and deep learning
  method.
\newblock In {\em Journal of Physics: Conference Series}, volume 1213, page
  022001. IOP Publishing, 2019.

\bibitem{park2008method}
Hojoon Park.
\newblock A method for controlling the mouse movement using a real time camera.
\newblock {\em Brown University, Providence, RI, USA, Department of computer
  science}, 2008.

\bibitem{noreen2015hand}
Uzma Noreen, Mutiullah Jamil, and Nazir Ahmad.
\newblock Hand detection using hsv model.
\newblock {\em Hand}, 6(12), 2015.

\bibitem{grif2018human}
Horatiu-Stefan Grif and Trian Turc.
\newblock Human hand gesture based system for mouse cursor control.
\newblock {\em Procedia Manufacturing}, 22:1038--1042, 2018.

\bibitem{golash2020economical}
Richa Golash and Yogendra~Kumar Jain.
\newblock Economical and user-friendly design of vision-based natural-user
  interface via dynamic hand gestures.
\newblock {\em International Journal of Advanced Research in Engineering and
  Technology}, 11(6), 2020.

\bibitem{globefire}
Globefire.
\newblock globefire/hand detection tracking opencv-.

\bibitem{yi2018long}
Chengming Yi, Liguang Zhou, Zhixiang Wang, Zhenglong Sun, and Changgeng Tan.
\newblock Long-range hand gesture recognition with joint ssd network.
\newblock In {\em 2018 IEEE International Conference on Robotics and
  Biomimetics (ROBIO)}, pages 1959--1963. IEEE, 2018.

\bibitem{liu2019hand}
Peng Liu, Xiangxiang Li, Haiting Cui, Shanshan Li, and Yafei Yuan.
\newblock Hand gesture recognition based on single-shot multibox detector deep
  learning.
\newblock {\em Mobile Information Systems}, 2019, 2019.

\bibitem{ni2018light}
Zihan Ni, Jia Chen, Nong Sang, Changxin Gao, and Leyuan Liu.
\newblock Light yolo for high-speed gesture recognition.
\newblock In {\em 2018 25th IEEE International Conference on Image Processing
  (ICIP)}, pages 3099--3103. IEEE, 2018.

\bibitem{tanmaiehand}
Upadrasta Tanmaie and Ch~Srinivasa Rao.
\newblock Hand posture detection and classification using you only look once
  (yolo v2) object detector.

\bibitem{haratiannejadi2019smart}
Kianoush Haratiannejadi, Neshat~Elhami Fard, and Rastko~R Selmic.
\newblock Smart glove and hand gesture-based control interface for multi-rotor
  aerial vehicles.
\newblock In {\em 2019 IEEE International Conference on Systems, Man and
  Cybernetics (SMC)}, pages 1956--1962. IEEE, 2019.

\bibitem{tan2019efficientnet}
Mingxing Tan and Quoc~V Le.
\newblock Efficientnet: Rethinking model scaling for convolutional neural
  networks.
\newblock {\em arXiv preprint arXiv:1905.11946}, 2019.

\bibitem{ashtari2020indoor}
Erfan Ashtari, Mohammad~Amin Basiri, Saeid~Mohammadi Nejati, Hemen Zandi,
  Seyyed Hossein~SeyyedAghaei Rezaei, Mehdi~Tale Masouleh, and Ahmad Kalhor.
\newblock Indoor and outdoor face recognition for social robot, sanbot robot as
  case study.
\newblock In {\em 2020 28th Iranian Conference on Electrical Engineering
  (ICEE)}, pages 1--7. IEEE, 2020.

\end{thebibliography}

\end{document}